\pdfoutput=1

\documentclass{article}

\usepackage{times}
\usepackage{graphics}
\usepackage{graphicx} 
\usepackage{subfigure} 

\usepackage{natbib}

\usepackage{algorithm}
\usepackage{algorithmic}

\usepackage{hyperref}

\usepackage[accepted]{icml2015}

\usepackage{amsmath}

\floatstyle{plain}
\newfloat{myalgo}{tbhp}{mya}

{\begin{myalgo}[#1]
\centering
\begin{minipage}{#2}
\begin{algorithm}[H]}%
{\end{algorithm}
\end{minipage}
\end{myalgo}}

\icmltitlerunning{Adaptive Scan Gibbs Sampler}

\begin{document} 

\twocolumn[
\icmltitle{Adaptive Scan Gibbs Sampler for \\ 
           Large Scale Inference Problems}

\icmlauthor{Vadim Smolyakov}{vss@csail.mit.edu}
\icmlauthor{Qiang Liu}{qliu@csail.mit.edu}
\icmlauthor{John W. Fisher III}{fisher@csail.mit.edu}
\icmladdress{CSAIL, MIT, Cambridge, MA 02139, USA}

\icmlkeywords{adaptive gibbs sampling, bayesian lasso, dirichlet process}

\vskip 0.3in
]

\begin{abstract} 
For large scale on-line inference problems the update strategy is critical for performance. We derive an adaptive scan Gibbs sampler that optimizes the update frequency by selecting an optimum mini-batch size. We demonstrate performance of our adaptive batch-size Gibbs sampler by comparing it against the collapsed Gibbs sampler for Bayesian Lasso, Dirchlet Process Mixture Models (DPMM) and Latent Dirichlet Allocation (LDA) graphical models.
\end{abstract} 

\section{Introduction}
\label{intro}

Big data problems present a computational challenge for iterative updates of global and local parameters. On-line algorithms that fit model parameters on a sub-sampled set of data provide a stochastic gradient approximation of the posterior \cite{hoffman2013jmlr}. We focus on Gibbs sampling algorithms that provide exact posterior computation in large scale inference problems.

On-line variational inference algorithms are fast but result in approximate posterior. MCMC algorithms on the other hand draw samples from the true posterior but take a long time to converge. Recent work on speeding up MCMC and improving variational approximation includes split-merge methods \cite{chang14hdp} and structured variational inference \cite{hoffman2014arxiv}. 

Adaptive MCMC algorithms \cite{Andrieu2008stats}, \cite{atchade2009tr}, \cite{roberts2009jcgs} have been developed to improve mixing and convergence rate by automotically adjusting MCMC parameters. The goal of this paper is to adapt the batch-size in order to improve asymptotic convergence of MCMC. Our algorithm resembles stochastic approximation found to be both theoretically valid and to work well in practice \cite{Mahendran2012aistats}. 

In this work, we show the importance of selecting the right batch size for big data inference problems. The main contribution of this paper is an adaptive batch-size Gibbs sampling algorithm that yields optimum update strategy demonstrated on Bayesian Lasso, Dirichlet Process Mixture Model (DPMM) and Latent Dirichlet Allocation (LDA) graphical models.

\section{Local and Global Updates}

\begin{figure}[t]
	\centering
	\includegraphics[width=0.2\textwidth]{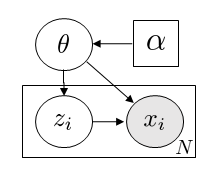}
	\caption{Graphical Model for Large Scale Inference Problems.}
    \label{fig:gm1}
\end{figure}

Consider a graphical model in Figure \ref{fig:gm1} where the observed data $x_i$ depends on both the global parameters $\theta$ and the local parameters $z_{i}$, where $i\in \{1,...,N\}$. The posterior distribution contains a product over a very large $N$:
\begin{equation}\label{equ:post1}
    P(\theta,z|x;\alpha) = P(\theta;\alpha)\prod_{n=1}^{N}P(z_{i}|\theta)P(x_i|z_i,\theta)
\end{equation}
We can re-write the posterior in (\ref{equ:post1}) as 
\begin{equation}\label{equ:post2}
    P(\theta,z) \propto \exp\{\psi(\theta) + \sum_i s(\theta, z_i)\} 
\end{equation}
where we choose exponential family representation for convenience. In the traditional Gibbs sampling algorithm, the full conditional updates alternate between:
\begin{eqnarray}\label{equ:post3}
    \{z_i\} &\sim& P(z_i|\theta) \propto \exp\{s(\theta,z_i)\} \\
    \theta &\sim& P(\theta|\{z_i\}) \propto \exp\{\psi(\theta) + \sum_{i=1}^{N} s(\theta,z_i)\} 
\end{eqnarray}
The global update of $\theta$ is inefficient since it requires an update of every $z_i$ and
 the summation over a very large $N$. Since the global parameters $\theta$ appear in each local Gibbs sampling update of $z_i \sim \exp\{s(\theta,z_i)\}$, we can achieve smaller variance by updating $\theta$ more often. Thus, we can find an optimum trade-off between the frequency of latent variable updates and Gibbs sampling performance.

\subsection{Optimizing Mini-Batch Size}

In evaluating MCMC performance we examine mixing time and the convergence rate to stationary distribution \cite{MurphyML}. Consider the following update frequencies after burnin for the graphical model in Figure \ref{fig:gm1}: 
\begin{align}\label{equ:minibatch1}
	\theta &\rightarrow z_1  \rightarrow \theta \rightarrow z_2 \rightarrow \theta \rightarrow z_3 \rightarrow ... \\
    \theta &\rightarrow z_1 \rightarrow z_2 \rightarrow \theta \rightarrow z_3 \rightarrow z_4 \rightarrow ... \label{equ:minibatch2} \\
    \theta &\rightarrow \theta \rightarrow z_1 \rightarrow \theta \rightarrow \theta \rightarrow z_2 \rightarrow ... \label{equ:minibatch3}
\end{align}
Frequent updates of $\theta$ in (\ref{equ:minibatch1}) increase the number of samples $n$ and help reduce the variance in (\ref{equ:mcmcvar}). On the other hand, a larger mini-batch size in (\ref{equ:minibatch2}) results in lower autocorrelation $\rho_t$ in (\ref{equ:rho}) and therefore greater information content per $\theta$-sample. Finally, a fractional mini-batch size in (\ref{equ:minibatch3}) updates $\theta$ multiple times before updating the local parameters $z_i$.   

Let $\bar{f}=\frac{1}{n}\sum_i f(\theta_i)$ be an average of MCMC samples of a function of the global parameter $f(\theta)$. Suppose we are interested in estimating $E[f(\theta)]$. A natural estimator is the sample mean $\hat{f}=\frac{1}{n}\sum_{i=1}^{n}E[f(\theta_i)]$. Under squared error loss, our objecitve is to minimize the variance:
\begin{eqnarray}\label{equ:mcmcvar}
   E[(\bar{f} - \hat{f})^{2}] &=& \frac{1}{n^2}E[\sum_{i=1}^{n}(f_i-\hat{f})^{2}] \nonumber \\
                              &+& \frac{1}{n^2}\sum_{s\neq t}E[(f_s-\hat{f})(f_t-\hat{f})] 
\end{eqnarray}
Alternatively, we can represent the variance of $\bar{f}$:
\begin{equation}\label{equ:mcmcvar2}
	\mathrm{VAR}[\bar{f}] \approx \frac{\sigma^2}{n}[1+2\sum_{t=1}^{\infty}\rho_t]
\end{equation}
where $\rho_t$ is the lag-$t$ autocorrelation:
\begin{eqnarray}\label{equ:rho}
   \rho_t=\frac{\frac{1}{n-t}\sum_{i=1}^{n-t}(f_i-\bar{f})(f_{i+t}-\bar{f})}{\frac{1}{n-1}\sum_{i=1}^{n}(f_i - \bar{f})^{2}}
   \end{eqnarray}
Let $\tau_{int} \doteq 1 + 2 \sum_{t=1}^{\infty}\rho_t$ be the \textit{integrated autocorrelation time} of the MCMC, then $\mathrm{VAR}(\bar{f})=\sigma^2\tau_{int}/n$. Therefore, $\tau_{int}$ is a measure of efficiency of the estimator of $\bar{f}$. In fact, $n_{eff}=n/\tau_{int}$ is the effective sample size. The autocorrelation function is commonly used to asses convergence and is therefore a good candidate for optimizing the batch size.

Consider a Gibbs sampling algorithm with mini-batch size $m$ for a graphical model in Figure \ref{fig:gm1}. Let $w_z$ and $w_{\theta}$ be the time of each local and global update, respectively. Given a fixed time budget $T$, the number of $\theta$-samples we can get is
\begin{equation}\label{equ:nsamples}
	n = \frac{T}{m w_z + w_{\theta}}
\end{equation}
Therefore, we can re-write the variance in (\ref{equ:mcmcvar2}) as
\begin{equation}\label{equ:mcmcvar3}
	\mathrm{VAR}[\bar{f}] \approx \frac{\sigma^2}{T}(mw_{z} + w_{\theta})\tau_{int}
\end{equation}
Because $\sigma^2$ and $T$ are fixed, we can formulate our mini-batch objective function as:
\begin{equation}\label{equ:obj}
	\min_m (mw_{z}+w_{\theta})\tau_{int}(m)
\end{equation}
Note as the mini-batch size $m$ decreases, the time cost $mw_z+w_{\theta}$ will decrease linearly, while $\rho_t$ and therefore $\tau_{int}$ will increase. This leads to Algorithm \ref{alg:minibatch} for selecting an optimum mini-batch size. 

\begin{algorithm}[tb]
\caption{Adaptive Batch Size Algorithm}
\label{alg:minibatch}
\begin{algorithmic}[1]
\STATE Define the mini-batch range $M=\{m_1,m_2,...,m_M\}$ and the number of samples $n$
\FOR{$m = m_1, m_2, ..., m_M$}
\STATE Run MCMC chain after burnin with batch size equal to $m$ for $n$ iterations 
\STATE Record $n$ samples of $\theta$ and compute $\tau_{int}$
\STATE Compute the objective: \\$f(m) = (mw_z+w_{\theta})\tau_{int}(m)$
\ENDFOR
\STATE Return $m^{*} = \arg \min_m f(m)$.
\end{algorithmic}
\end{algorithm}
The adaptive batch size algorithm consists of two phases: adaptation and sampling. During adaptation phase, the optimum mini-batch size is selected after burnin in $O(M(n+t_{max}))$ time and $O(Mn(|\theta|+t_{max}))$ space, where $t_{max}$ is the max autocorrelation lag. During the sampling phase, the optimimum mini-batch size is used to achieve fast convergence. 


\section{Experimental Results}

To evaluate performance of the adaptive batch size Gibbs sampler we consider Bayesian lasso and Dirichlet Process Gaussian Mixture Model (DPGMM) graphical models. 

\subsection{Probit Regression with Non-conjugate Prior}

\begin{figure}[t]
	\centering
    \includegraphics[width=0.2\textwidth, trim={10 10 10 10}]{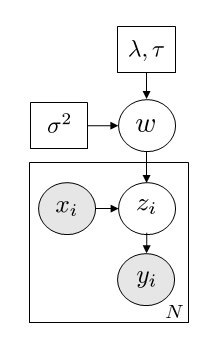}
	\caption{Graphical Model for Bayesian lasso.}
    \label{fig:gm2}
\end{figure}

Consider the model
\begin{equation}\label{equ:probit1}
  y|X,w,\sigma^2 \sim N(Xw,\sigma^2 I_n)
\end{equation}
where $y\in R^n$ is the response vector, $X$ is a known $n\times d$ design matrix, and $w \in R^d$ is the unknown vector of regression coefficients. Alternatively, we can re-write the expression as:
\begin{equation}\label{equ:probit2}
  w = \arg \min_{w \in R^d}(||y-Xw||_{2}^{2} + \lambda ||w||_1)
\end{equation}
Note that $w$ as defined in (\ref{equ:probit2}) can be veiwed as a posterior mean of $w$ with likelihood (\ref{equ:probit1}) and prior:
\begin{equation}\label{equ:probit3}
 w_i | \sigma^2 \sim \mathrm{iid}~\mathrm{Laplace}(\lambda/\sigma),~i \in \{1,...,d\}
\end{equation}
where $\mathrm{Laplace}(\alpha)$ distribution has density $f(u) = (\alpha/2)\exp(-\alpha |u|)$. The Laplace prior in (\ref{equ:probit3}) can be represented as a scale mixture of normal distributions \cite{park2008jasa}:
\begin{equation}\label{equ:probit4}
 w | \sigma^2,\tau \sim N(0,\sigma^2 D_{\tau}),~\tau_i \sim \mathrm{Exp}(\lambda^2/2)
\end{equation}
where $D_{\tau}=\mathrm{Diag}(\tau_1,...,\tau_d)$.
This frameworks is known as Bayesian lasso. Figure \ref{fig:gm2} shows the graphical model for Bayesian lasso. The generative model is defined as follows:
\begin{eqnarray}\label{equ:probit3}
  w &\sim&  N(0,\sigma^2 D_{\tau}), ~\tau_i = (\lambda^2/2)e^{-\lambda^2 \tau_i^{2}/2}\\
  z_i &\sim& N(w^{T}x_i,\sigma^2 I_{d}), ~y_i = \mathrm{sgn}(z_i)
\end{eqnarray}
To reduce the dimensionality of $w$ for big data applications, a one-sided Laplacian or an exponential prior is introduced on the covariance of the hyperplane $w$. However, this prior is non-conjugate and requires an efficient sampling algorithm. Conditioned on prior hyper-parameters, the label updates $z_i$ are identical to Algorithm (\ref{alg:probit}). The hyper-parameter are updated as derived in \cite{park2008jasa}:
\begin{equation}\label{equ:probit5}
   \sigma^{2} \sim \mathrm{I}\Gamma((n-1)/2+d/2,\frac{1}{2}||y-Xw||_{2}^{2}+\frac{1}{2}w'D^{-1}_{\tau}w)
\end{equation}
\begin{equation}\label{equ:probit6}
   1/\tau_j^2 \sim \mathrm{IN}(\sqrt{\frac{\lambda^2\sigma^2}{w_j^2}},\lambda^2) \\
\end{equation}
\begin{equation}\label{equ:probit7}
   w \sim \mathrm{N}(A^{-1}X^{T}y, \sigma^2 A^{-1})
\end{equation}
where $A = X^{T}X+D_{\tau}^{-1}$. The full conditional updates above when combined with incremental updates lead to the following incremental Gibbs sampling Algorithm \ref{alg:probit}.

\begin{algorithm}
\caption{Incremental Gibbs Sampler for Bayesian Lasso}
\label{alg:probit}
\begin{algorithmic}[1]
\STATE Init $z_i~\forall i \in \{1,N\}$ and sample a mini-batch $i \sim \mathrm{Unif}[1,N]$
\STATE Local update: $z_{i}|w \sim N(w^{T}x_i,1)~1(z_i \geq 0)$ if $y_i \geq 0$, and $N(w^{T}x_i,1)~1(z_i \leq 0)$, otherwise. 
\STATE Global update: $w$, $\sigma^2$, $\tau^2$ as in (\ref{equ:probit5})-(\ref{equ:probit7})
\end{algorithmic}
\end{algorithm}
The mini-batch Gibbs sampler updates are analogous to stochastic gradient and are expected to give a computational advantage for big data applications with a limited time budget $T$.\\

\begin{figure}[t]
	\centering
    \includegraphics[width=0.5\textwidth, trim={10 10 10 10}]{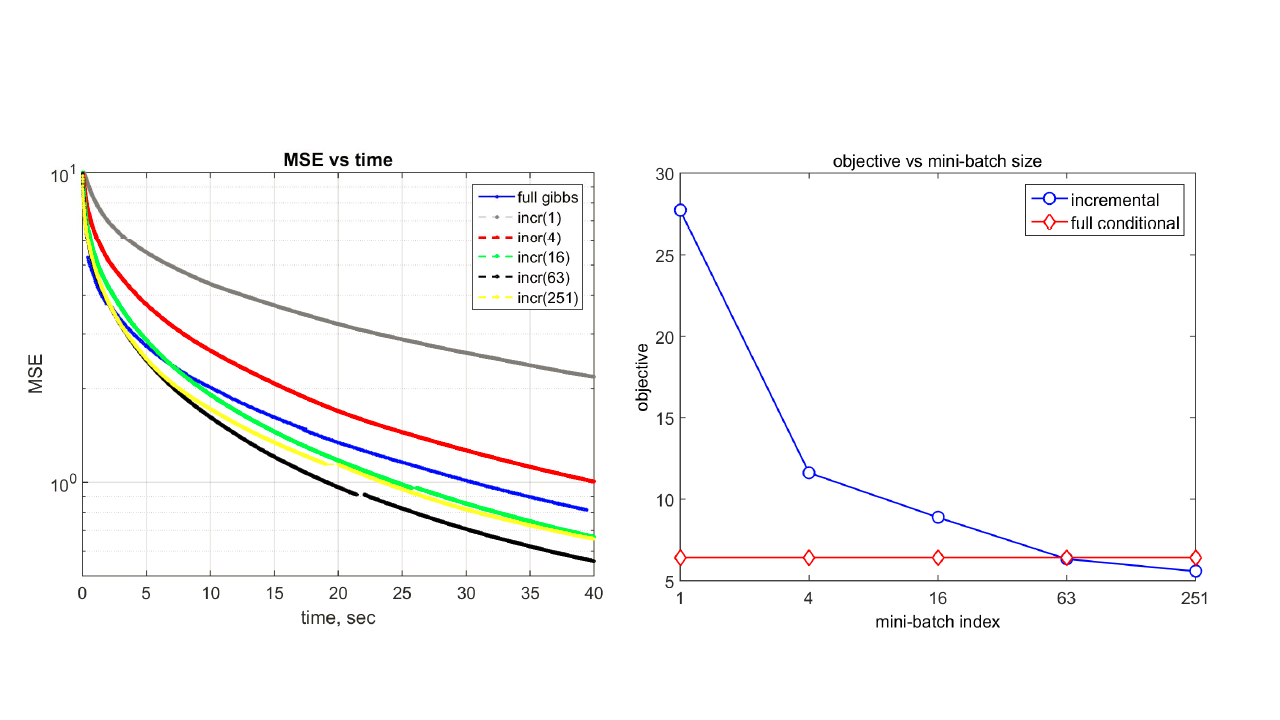}
	\caption{MSE vs time plot for sequential scan (left) and corresponding objective function (right).}
	\label{fig:probit2}
\end{figure}

\subsubsection{Experimental Results}

The experimental results were obtained by averaging samples of three parallel MCMC chains with Estimated Potential Scale Reduction (EPSR) used as a convergence criterion:
\begin{equation}
   \hat{R} = \sqrt{\frac{\mathrm{var(\psi|y)}}{W}}
\end{equation}
where $\mathrm{var}(\psi|y) = \frac{n-1}{n}W + \frac{1}{n}B$, with $B$ and $W$ representing the in-between chain and within-chain variance, respectively \cite{BDA3}.

Figure \ref{fig:probit2} shows the MSE for a dataset consisting of $n=1.2K$ points in $R^{4}$ averaged over multiple trials. We can see that the lowest MSE for a fixed time budget is achieved by mini-batch sizes $63$ and $251$ outperforming the full-conditional Bayesian lasso Gibbs sampler. Similarly, the objective plot shows that mini-batch size $251$ will be selected during the sampling phase. Thus, the adaptive batch-size Gibbs sampler first cycles through mini-batch sizes $M=\{1,4,16,63,251\}$ during the adaptation phase and fixes the mini-batch size to $251$ during the sampling phase.



\subsection{Dirichlet Process Mixture Model}

Dirichlet Process Mixture Model (DPMM) uses non-parametric priors for modeling mixtures with infinite number of clusters:
\begin{equation}
   G(\theta) = \sum_{k=1}^{\infty}\pi_k \delta_{\theta_k}(\theta)
\end{equation}
where $\pi_k \sim \mathrm{GEM}(\alpha)$ and $\theta_k \sim H(\lambda)$.
A Dirichlet process is a distribution over probability measures $G:\Theta \rightarrow R^{+}$ and it defines a conjugate prior for arbitrary measurable spaces:
\begin{equation}
   G|\bar{\theta}_1,...,\bar{\theta}_N,\alpha, H  \sim \mathrm{DP}(\alpha + N, \frac{1}{\alpha+N}(\alpha H + \sum_{i=1}^{N}\delta_{\theta_i})) 
\end{equation}
Applied to mixture modeling, we can describe the graphical model for DPMM in Figure \ref{fig:gm3} as follows:
\begin{eqnarray}
   \pi &\sim& GEM(\alpha) \\
   z_i &\sim& \pi \\
   \theta_k &\sim& H(\lambda) \\
   x_i &\sim& F(\theta_{z_i})
\end{eqnarray}

\begin{figure}[t]
	\centering
    \includegraphics[width=0.2\textwidth, trim={10 10 10 10}]{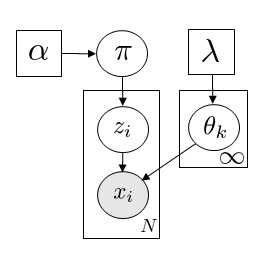}
	\caption{Graphical Models for DPMM.}
    \label{fig:gm3}
\end{figure}

The simplest way to fit a DPMM is to modify the collapsed Gibbs sampler for a finite mixture model. By exchangeability, we can assume that $z_i$ is the most recent assignment and therefore,
\begin{equation}
 p(z_i|z_{-i}, \alpha) = \frac{1}{\alpha+N-1}(\alpha 1_{z_i = k^{*}} + \sum_{k=1}^{K}N_{k,-i}1_{z_i=k})
\end{equation}
The rest of the algorithm can be summarized as follows:

\begin{algorithm}
\caption{Mini-Batch Gibbs Sampler for DPMM}
\label{alg:probit}
\begin{algorithmic}[1]
\STATE Init $z_i~\forall i \in \{1,N\}$ and select optimum mini-batch as in Algorithm 1
\STATE Local Update: Compute $p(z_i=k|z_{-i},D) = \frac{N_{k,-i}}{\alpha+N-1}p(x_i|x_{k\setminus i})$
and $p(z_i=new|z_{-i},D) = \frac{\alpha}{\alpha+N-1}p(x_i|x_{k\setminus i})$. Sample $z_i \sim p(z_i|\cdot)$
\STATE Global Update: $\mu_k$ and $\Sigma_k$ based on the new $z_i$.
\end{algorithmic}
\end{algorithm}

\subsubsection{Experimental Results}

\begin{figure}[t]
	\centering
	\vspace{-5pt}
    \includegraphics[width=0.5\textwidth, trim={10 10 10 10}]{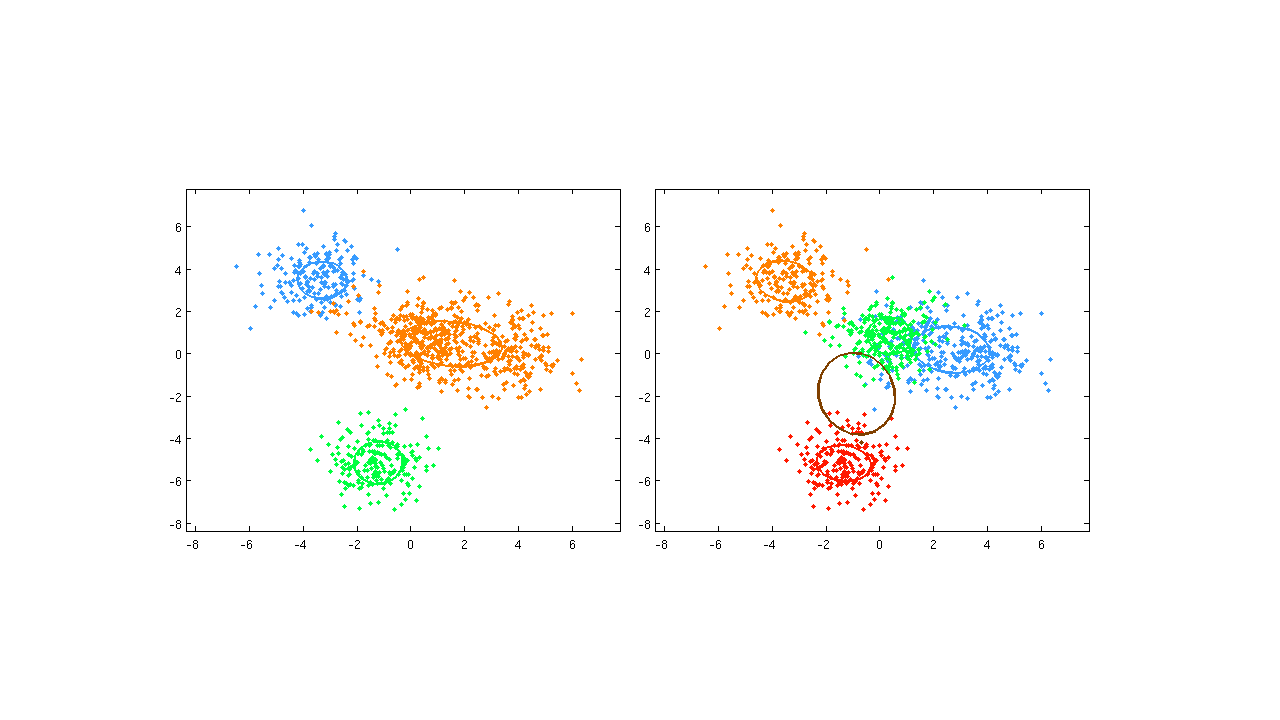}
    \vspace{-20pt}
	\caption{DPMM Posterior Clustering for Collapsed Gibbs (left) and Mini-Batch Gibbs (right).}
    \label{fig:probit3}
\end{figure}

Figure \ref{fig:probit3} shows the posterior DPMM clustering of $n=1K$ data points in $R^{2}$ with concentration parameter $\alpha=1$ and the true number of clusters set to $5$. We can see that the mini-batch Gibbs samplers produced the true number of clusters by iteration $200$ in comparison with the collapsed Gibbs sampler. 

\begin{figure}[t]
	\centering
	\vspace{-5pt}
    \includegraphics[width=0.5\textwidth, trim={10 10 10 10}]{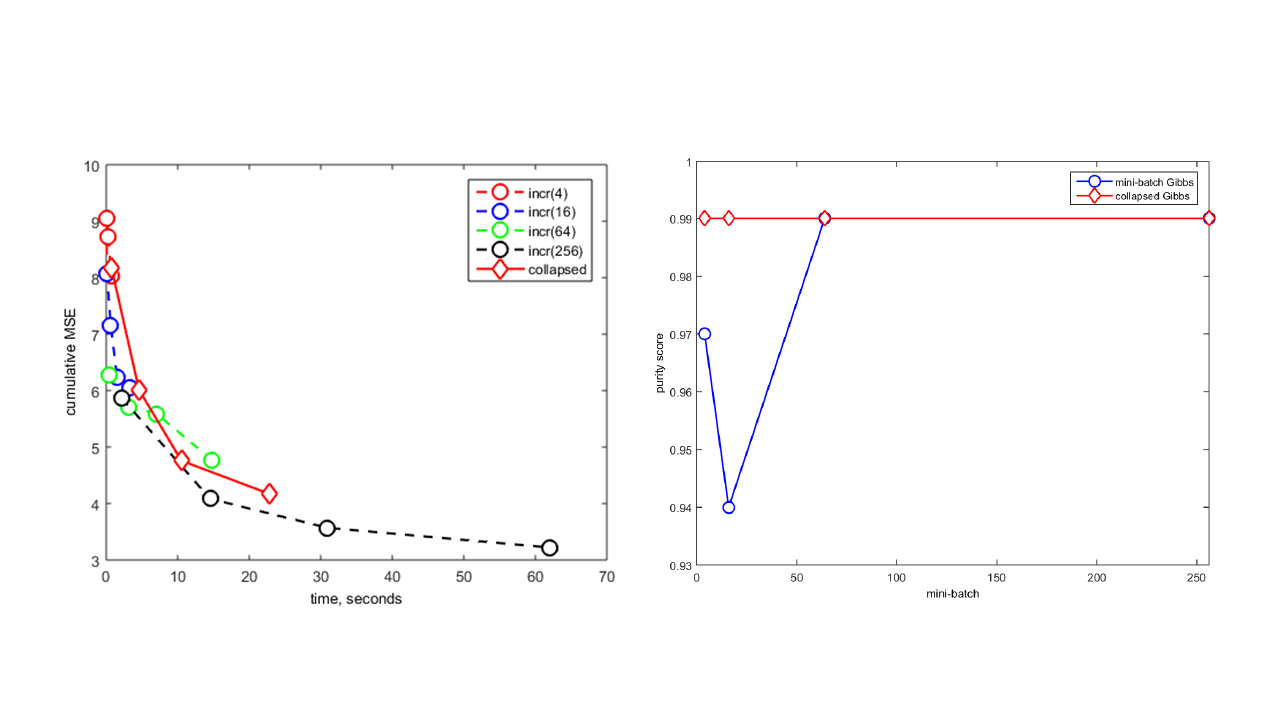}
    \vspace{-5pt}
	\caption{DPMM MSE (left) and purity scores (right).}
    \label{fig:probit4}
\end{figure}

Figure \ref{fig:probit4} shows the MSE and purity scores computed at $N=200$ iterations comparing the Gibbs sampling posterior clustering with the ground truth. The MSE scores were computed by matching inferred clusters using the Hungarian algorithm with Euclidean distance as the cost matrix. The purity score is defined as $\sum_i \frac{N_i}{N}p_{i}$, where $N_i = \sum_{j=1}^{C}N_{ij}$ and $p_i = \max_j p_{ij}$, with $p_{ij} = N_{ij}/N_{i}$ where $N_{ij}$ is defined as the number of objects in cluster $i$ that belongs to class $j$.

\subsection{Latent Dirichlet Allocation}

Latent Dirichlet Allocation (LDA) is a generative ad-mixture topic model in which every word is assigned to its own topic $z_{id} \in \{1,...,K\}$ drawn from a document specific distribution $\theta_d$ \cite{blei2003jmlr}.

\begin{figure}[t]
	\centering
    \includegraphics[width=0.3\textwidth, trim={10 10 10 10}]{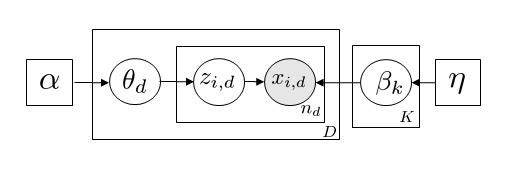}
	\caption{Latent Dirichlet Allocation (LDA) graphical model.}
    \label{fig:gm3}
\end{figure}

The generative LDA model can be described as follows:
\begin{eqnarray}
   \theta_d | \alpha &\sim& \mathrm{Dir}(\alpha_1,...,\alpha_K) \\
   z_{id} | \theta_d &\sim& \mathrm{Cat}(\theta_d) \\
   \beta | \eta &\sim& \mathrm{Dir}(\eta_1,...,\eta_V) \\
   x_{id}|z_{id} = k, \beta &\sim& \mathrm{Cat}(\beta_k)
\end{eqnarray}

It is straightforward to derive a full conditional Gibbs sampling algorithm for LDA. However, one can get better performance by analytically integrating out $\theta_d$ and $\beta$. This leads to the following collapsed Gibbs sampler expression:
\begin{equation}
	p(z_i = j|z_{-i},w) \propto \frac{n_{-i,j}^{(w_i)}+\beta_i}{n_{-i,j}^{(\cdot)}+\sum_i \beta_i} \frac{n_{-i,j}^{(d_i)}+\alpha_j}{n_{-i,\cdot}^{(d_i)}+\sum_j \alpha_j}
\end{equation}
where $n_{-i,j}^{(w_i)}$ is the number of words that belong to topic $j$ in the entire corpus excluding the $i$-th word and $n_{-i,j}^{(d_i)}$ is the number of words assigned to topic $j$ in document $d_{i}$ excluding the $i$-th word.

Rather than waiting for the collapsed Gibbs sampler to iterate over all documents $D$ before updating the global topic parameter $\beta$, we can divide the corpus into a set of mini-batches $M \in \{m_1,...,m_M\}$. This leads to the following mini-batch Gibbs sampling algorithm:

\begin{algorithm}
\caption{Mini-Batch Gibbs Sampler for LDA}
\label{alg:probit}
\begin{algorithmic}[1]
\STATE Init $z_i~\forall i \in \{1,N\}$ and sample a mini-batch $i \sim \mathrm{Unif}[1,N]$
\STATE Local Update: Compute $p(z_i=k|z_{-i},x_{id})$, update $\theta_d$ and sample $z_i$
\STATE Global Update: $\beta_k$ based on the new $z_i$:
\STATE $p(\beta_k|\cdot) = \mathrm{Dir}(\eta_v +\sum_i\sum_d 1(x_{id}=v, z_{id}=k))$
\end{algorithmic}
\end{algorithm}

\subsubsection{Experimental Results}

\begin{figure}[t]
	\centering
	\vspace{-5pt}
    \includegraphics[width=0.5\textwidth, trim={10 10 10 10}]{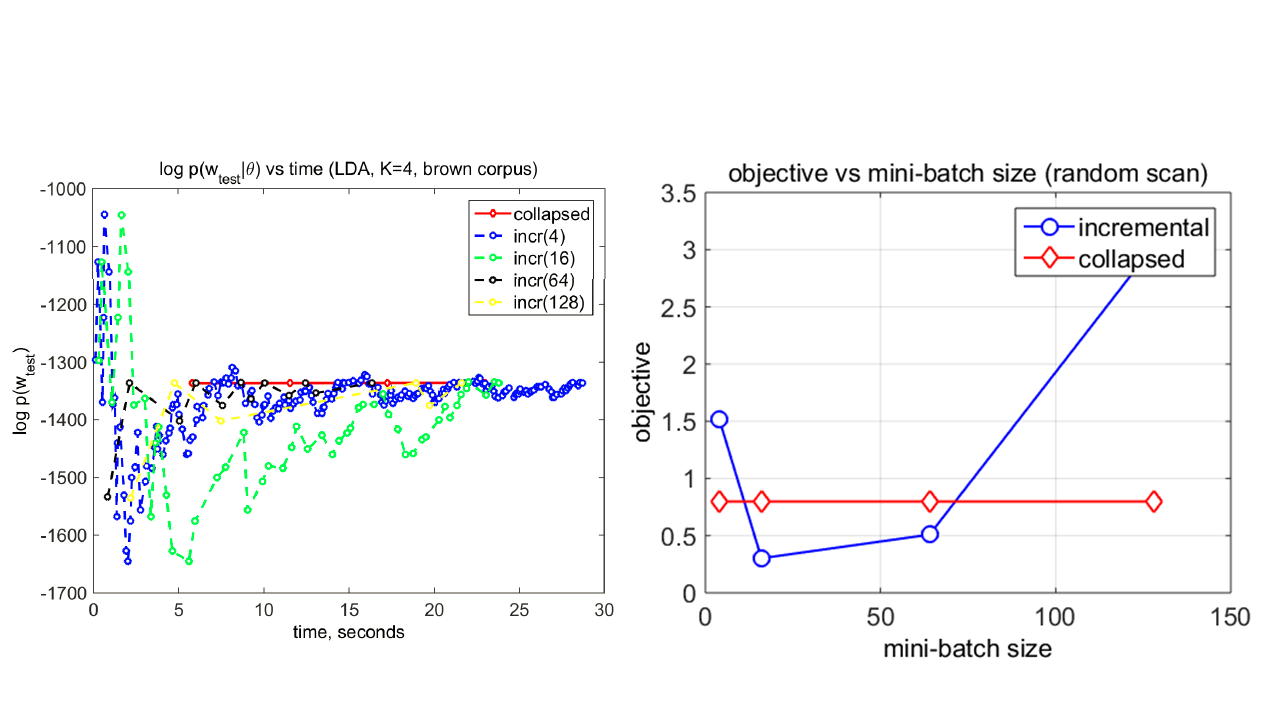}
    \vspace{-5pt}
	\caption{LDA perplexity (left) and objective function (right).}
    \label{fig:lda1}
\end{figure}

Figure \ref{fig:lda1} shows the experimental results on the Brown corpus with $K=4$ topics, $V=6K$ dictionary and $D=250$ documents. The perplexity of the test set was to evaluate performance:
\begin{equation}
  \mathrm{Perplexity}(w_{test}) = \exp\{-\frac{1}{D_{test}}\sum_d \frac{1}{n_d}\sum_{w \in n_d}\log p(w_{test})\}
\end{equation}

Perplexity is computed by evaluating the expression below over $S$ independent chains:
\begin{align}\label{equ:perplexity4}
   \log p(w_{test}) = \sum_{j,w} N_{j,w}^{test} \log \frac{1}{S} \sum_s \sum_k \theta_{k|j}^{s} \phi_{w|k}^{s} \\     
 \mathrm{where}~\theta_{k|j}^{s} = \frac{\alpha + N_{kj}^{s}}{\sum_k \alpha_k + N_{j}^{s}},~\phi_{w|k}^{s}=\frac{\beta+N_{wk}^{s}}{\sum_w \beta_w + N_{k}^{s}}
\end{align}
 where $N_{kj}^{s}$ is the number of words assigned to topic $k$ in document $j$ and $N_{wk}^{s}$ is the number of words assigned to topic $k$ across the entire corpus. Alternatively, $\log p(w_{test})$ probability can be computed using chib-IS estimator and other methods described in \cite{wallach2009icml}.

The objective function shows a preference for mini-batch size $16$ which also leads to smaller perplexity for a fixed time period.

\section{Discussion}

\subsection{Mini-Batch Gibbs Sampler}

The mini-batch Gibbs sampler uses a sequential sampling scheme alternating between sampling of the global parameters $\theta$ and local parameters $z_i$. The optimum sampling frequency is selected by optimizing the mini-batch objective function $f(m) = (mw_z + w_{\theta})\tau_{int}(m)$. MSE gain is achieved by updating the global parameters more frequently in comparison to the collapsed Gibbs sampler. This is evident for graphical models with a hierarchical structure such as Bayesian Lasso, DPMM and LDA.\\

The mini-batch Gibbs sampler is well suited for sampling algorithms used in large scale inference settings such as Stochastic Gradient Descent (SGD) \cite{li2014kdd}. If the mini-batchsize is set to $m=1$, this is the case of standard SGD, if $m=N$ we have the collapsed Gibbs sampler. Thus, we can achieve better MSE by choosing $1\leq m \leq N$ according to the mini-batch objective function. \\

The intuition for this is that one can get a fairly good initial estimate of the local hidden variables knowing the global parameters by evaluating just a few data points. Thus having a noisy estimate of global parameters enables rapid movement through the parameter space in a Gibbs sampling framework. In addition to the improvements in speed, mini-batch Gibbs sampler is less likely to get stuck in local minima due to a certain amount of noise added to parameter estimates.\\ 

The mini-batch Gibbs sampler algorithm consists of two phases: adaptation and sampling. During the adaptation phase, the optimum mini-batch size is chosen by a random shuffle through a fixed range of mini-batch sizes $m=\{1,...,M\}$. After which the optimum mini-batch is used for sampling during the sampling phase. While the algorithm introduces a time overhead of $M\times n$ samples, the chain continues to mix during the adaptation phase. To reduce the overhead, a logarithmic search of mini-batch size was used to minimize $M$. \\

\begin{figure}[t]
	\centering
	\vspace{10pt}
    \includegraphics[width=0.3\textwidth, trim={10 10 10 10}]{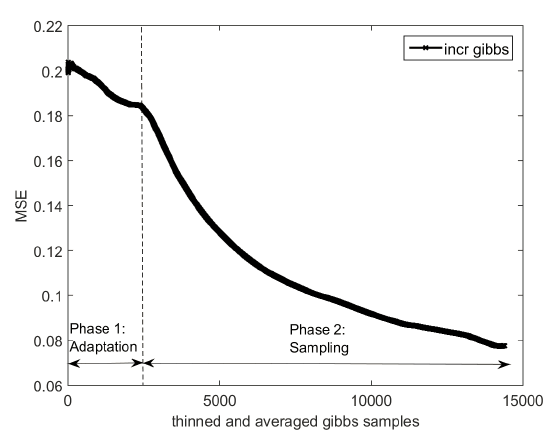}
	\caption{MSE vs time adaptation and sampling phase plot.}
    \label{fig:disc1}
\end{figure}

\begin{figure}[t]
	\centering
    \includegraphics[width=0.5\textwidth, trim={10 10 10 10}]{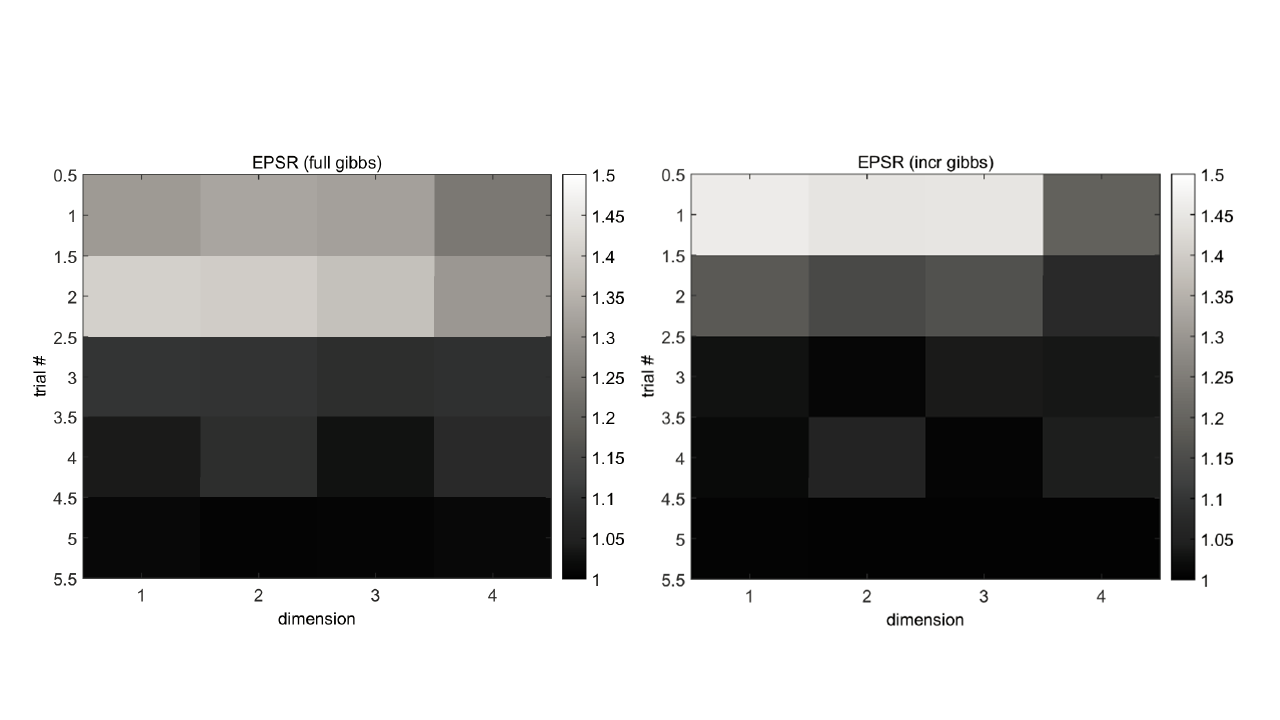}
	\caption{EPSR plot for full Gibbs sampler (left) and optimum mini-batch Gibbs sampler (right).}
    \label{fig:disc2}
\end{figure}

In addition, the mini-batch objective was selected to use commonly used MCMC diagnostic functions that present little computational overhead such as the integrated autocorrelation $\tau_{int}$ and empirical sampling time $w_z$ and $w_{\theta}$. Figure \ref{fig:disc1} illustrates the two phases of the mini-batch Gibbs sampling algorithm for Bayesian Lasso. Also shown in Figure \ref{fig:disc2} are the EPSR convergence metrics comparing a full conditional Gibbs sampler (left) and the collapsed Gibbs sampler with optimum mini-batch size (right) for different number of trials and dimensions of the global parameter $w$.\\

Notice that the adaptation phase acts as initialization for the sampling phase. In general, the mini-batch Gibbs sampling algorithm can be used to initialize batch optimization methods that converge faster in the vicinity of potentially global optimum.

\section{Conclusion}

We developed a mini-batch Gibbs sampling algorithm that outperformed state of the art collapsed Gibbs sampler for hierarchical graphical models in large scale inference setting. The complexity of the algorithm is negligible compared to reduction in asymptotic variance as a result of optimum mini-batch selection. We demonstrated the performance of the algorithm on Bayesian Lasso, DPMM, and LDA graphical models.

\bibliography{svi}
\bibliographystyle{icml2015}

\end{document}